\documentclass[oribibl, 12pt]{llncs}
\pdfoutput=1
\usepackage{url, listings, subfigure} 
\usepackage[utf8]{inputenc}
\usepackage[flushleft,alwaysadjust]{paralist}
\usepackage[hyperfootnotes=false, pagebackref=true, citebordercolor={1 1 1}, linkbordercolor={0 0 1}, urlbordercolor={1 1 1}]{hyperref}
\usepackage[]{epsfig}
\usepackage[sectionbib]{natbib}
\bibpunct[:]{(}{)}{;}{a}{}{,}
\usepackage{graphics,latexsym}
\usepackage[inner=4cm,outer=4cm,top=4.5cm,bottom=5cm]{geometry}
\usepackage[margin=0pt,font=normalsize,labelfont=bf, format=hang]{caption}
\title{Computational Representation of Linguistic Structures using Domain-Specific Languages} 
\author{Fabian Steeg\inst{1}, Christoph Benden\inst{2} \& Paul O. Samuelsdorff\inst{3}} 
\institute{Computer Science for the Humanities, University of Cologne \and German Institute of Medical Documentation and Information, Cologne \and General Linguistics, University of Cologne}

\begin{document}
\maketitle
\thispagestyle{plain}
\begin{center}\today\end{center}
\begin{abstract}
We describe a modular system for generating sentences from formal definitions of underlying linguistic structures using do\-main-spe\-cif\-ic languages. The system uses Java in general, Prolog for lexical entries and custom domain-specific languages based on Functional Grammar and Functional Discourse Grammar notation, implemented using the ANTLR parser generator. We show how linguistic and technological parts can be brought together in a natural language processing system and how domain-specific languages can be used as a tool for consistent formal notation in linguistic description.
\end{abstract}

\section{Motivation and Overview}
This paper describes a system for generating sentences using domain-specific languages (DSL; see section \ref{dsls}) for the formal representation of underlying linguistic structures and lexical entries.\footnote{The described implementation and infrastructure for collaborative development are available online  (\url{http://fgram.sourceforge.net}).} The DSL implemented for underlying structures is based on representations in Functional Grammar (FG; \citealt{Dik1997a}). The grammar module and the lexicon are based on a revised and extended version of the implementation described in \cite{Samuelsdorff1989}. To evaluate the flexibility of our approach, we also implemented domain-specific languages for formal representations in Functional Discourse Grammer (FDG), which as FG explicitly demands ``formal rigor'' \citep[668]{HengeveldAndMackenzie2006}. Creating a computational implementation is a valuable evaluation tool for linguistic theories in general (cf. \citealt[4]{Bakker1994}). By actually generating linguistic expressions from representations used in a linguistic theory, an implementation can be used to evaluate and improve representational aspects of the theory.

\section{System Architecture}

\begin{figure}
\begin{center}
\includegraphics[width=14cm]{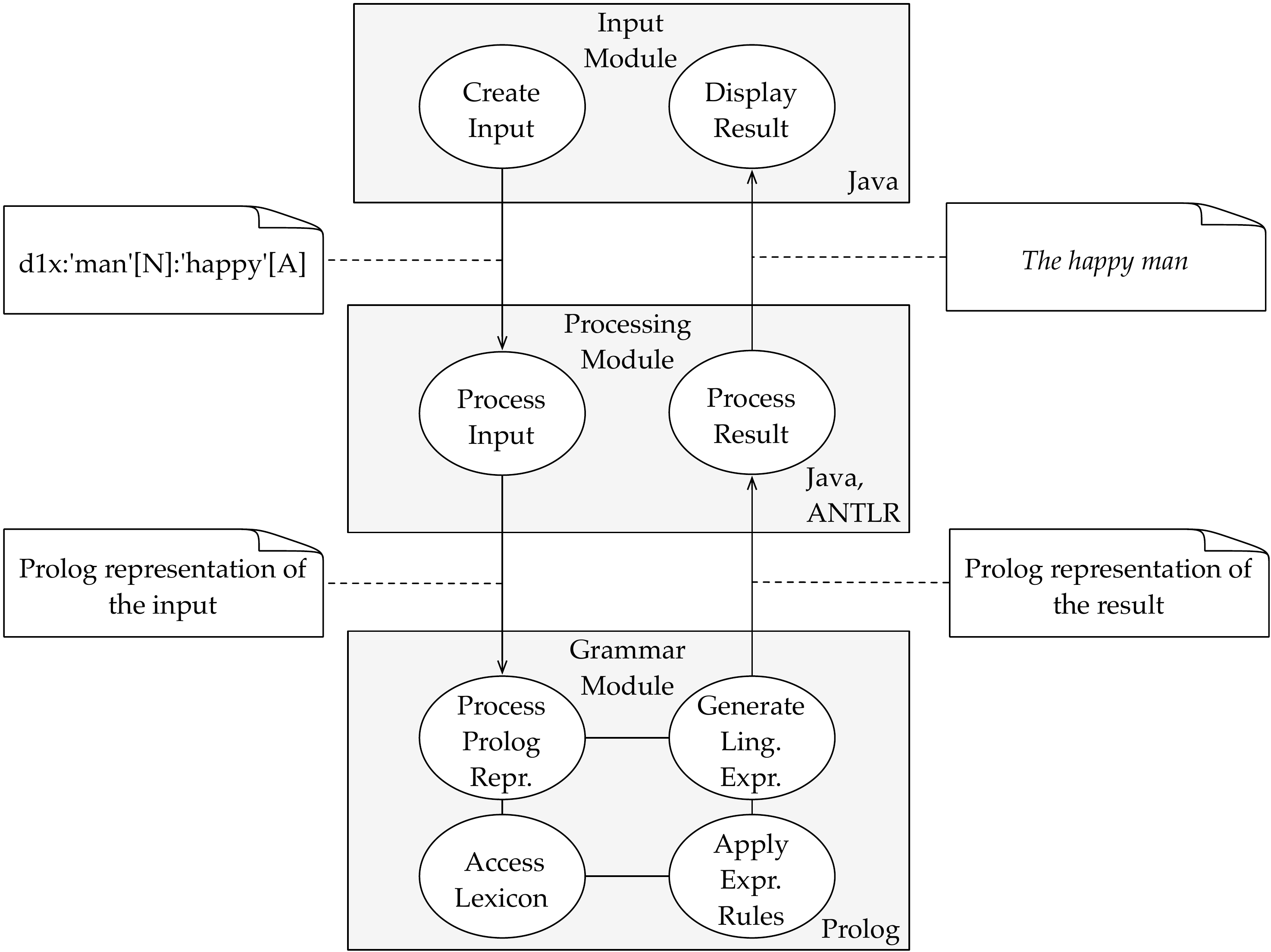}
\end{center}
\caption{System architecture} \label{sysflow}
\end{figure}

The system consists of individual, exchangeable modules for creating an underlying structure, processing that input and generating a linguistic expression from the input (cf. Fig.~\ref{sysflow} for an overview of the system architecture). In the \emph{input module} an underlying structure is created, edited and evaluated. The input is sent to the \emph{processing module}, which communicates with the \emph{grammar module}. When the generation is done, the user interface displays either the result of the evaluation, namely the linguistic expression generated from the input, or an error message (cf. Fig. \ref{screen} for sample output of the console-based implementation of the input module). The system architecture can be characterized as a three-tier architecture \citep{Eckerson1995}.

Such a modular approach has two main advantages. First, modules can be exchanged; for instance the \emph{input module} is implemented both as a desktop application and as a web-based user interface with the actual processing happening on a server (implemented using Java Server Pages on a Tomcat servlet container). Second, individual modules of our system can be combined with other natural language processing (NLP) components and so be reused in new contexts.

\section{Domain-Specific Languages}\label{dsls}

The usage of languages which are tailored for a specific domain (domain-specific languages, DSL) has a long tradition in computing (e.g. for configuration files) and has been acknowledged as a best practice in recent years (cf. \citealt{HuntAndThomas1999}). Domain-specific languages are also a central aspect of a  programming paradigm called language-oriented programming (cf. \citealt{Ward2003}). 

Our system uses Java as a general-purpose language, Prolog as a DSL for lexical entries and expression rules (see section \ref{lexical}, cf. \citealt{Macks2002} for a similar usage of Prolog), and a custom DSL for describing underlying structures, implemented using ANTLR, a tool for defining and processing domain-specific languages (\citealt{Parr2007}, \url{http://www.antlr.org/}). While e.g. in the domain of banking a DSL might describe credit rules, a linguist working with a model like FDG uses a DSL for linguistic description, in particular for the formal notation of underlying linguistic structures. With ANTLR, the form of the DSL is defined using a notation based on the \emph{Extended Backus-Naur Form} (EBNF, cf. \citealt{Wirth1977}, see Fig. \ref{antlr-rl-parser} for the format used by ANTLR). From that grammar definition a Java parser that can process the DSL is automatically generated by ANTLR.

\section{Linguistic Structures}

\subsection{Structures in Functional Grammar}\label{fg}

The \emph{processing module's} input format is a representation of the linguistic expression to be generated (cf. Fig. \ref{antlr-input} and \ref{screen}); its form is based on the representation of underlying structures given in \cite{Dik1997a}. The \emph{processing module} parses the input entered by the user and creates an internal object representation (cf. Fig. \ref{uml-tree}). This is then converted into the output format of the \emph{processing module}, a Prolog representation of the input (cf. Fig. \ref{prolog-ucs}), which is used by the \emph{grammar module} (cf. section \ref{lexical}). The mapping of the values used in the Prolog representation to those used in the input structure (like \emph{m} to \emph{plural}) is done in a Java properties file and therefore allows for configuration of the formal aspects of the input (which uses e.g. \emph{m}) independently of the implementation code that generates the expression (which uses e.g. \emph{plural}).

\begin{figure}
\begin{center}
\begin{verbatim}
(Past e:
    (d1x:'man'[N]:
        (Past Pf e:'give'[V]
            (d1x:'mary'[N])Ag  
            (dmx:'book'[N]:'old'[A])Go
            (x:'man'[N])RecSubj
        )
    )
    (d1x:'john'[N])0
)
\end{verbatim}
\caption{A nested underlying structure in Functional Grammar based on \cite{Dik1997a}, which is parsable by the generated ANTLR v2 parser (represents \emph{John is the man who was given the old book by Mary})}\label{antlr-input}
\end{center}
\end{figure}

\begin{figure}
\begin{verbatim}
>> (e:'love'[V]:(x:'man'[N])AgSubj (x:'woman'[N])GoObj)
The man loves the woman

>> (Past pf e:'give'[V]:
        (dmx:'farmer'[N]:'old'[A])AgSubj
        (imx:'duckling'[N]:'soft'[A])GoObj 
        (dmx:'woman'[N]:'young'[A])Rec)
The old farmers had given soft ducklings to the young women
\end{verbatim}
\caption{Sample output of the console-based implementation of the input module: a linguistic structure conforming to Functional Grammar notation is entered at the prompt ($\gg$), for which the linguistic expression is generated using the linguistic knowledge in the grammar module} \label{screen}
\end{figure}

\begin{figure}
\begin{center}
\mbox{\includegraphics[height=5cm]{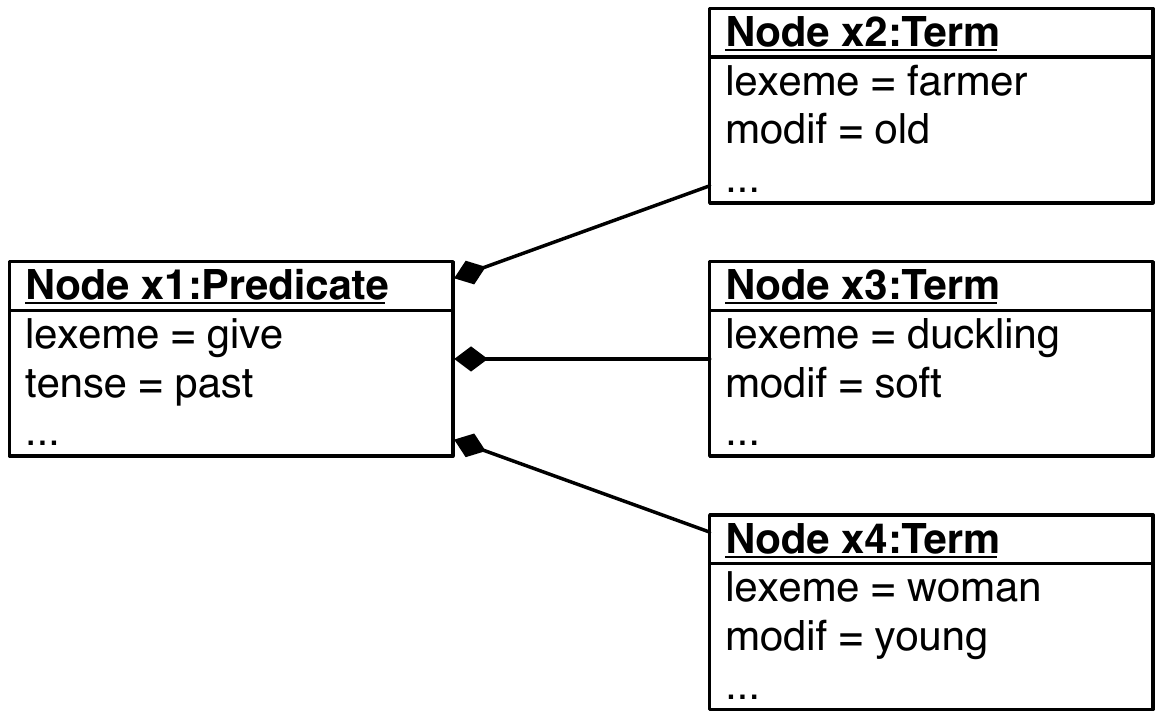}}
\caption{Internal representation of the second structure in Fig. \ref{screen} (represents \emph{The old farmers had given soft ducklings to the young women}): a tree of Java objects (in UML notation)}
\label{uml-tree}
\end{center}
\end{figure}

\begin{figure}
\begin{minipage}[hbt]{6cm}
\centering
\begin{verbatim}
node(x1, 0). node(x2, 1).
node(x3, 1). node(x4, 1).

prop(x1, type, pred).
prop(x1, tense, past).
prop(x1, perfect, true).
prop(x1, progressive, false).
prop(x1, mode, ind).
prop(x1, voice, active).
prop(x1, subnodes, [x2, x3, x4]).
prop(x1, lex, 'give').
prop(x1, nav, [V]).
prop(x1, det, def).

prop(x2, type, term).
prop(x2, role, agent).
prop(x2, relation, subject).
prop(x2, proper, false).
prop(x2, pragmatic, null).
prop(x2, num, plural).
prop(x2, modifs, [old]).
prop(x2, lex, 'farmer').
prop(x2, nav, [N]).
prop(x2, det, def).
\end{verbatim}
\end{minipage}
\hfill
\begin{minipage}[hbt]{6cm}
\centering
\begin{verbatim}
prop(clause, illocution, decl).
prop(clause, type, mainclause).

prop(x3, type, term).
prop(x3, role, goal).
prop(x3, relation, object).
prop(x3, proper, false).
prop(x3, pragmatic, null).
prop(x3, num, plural).
prop(x3, modifs, [soft]).
prop(x3, lex, 'duckling').
prop(x3, nav, [N]).
prop(x3, det, indef).

prop(x4, type, term).
prop(x4, role, recipient).
prop(x4, relation, restarg).
prop(x4, proper, false).
prop(x4, pragmatic, null).
prop(x4, num, plural).
prop(x4, modifs, [young]).
prop(x4, lex, 'woman').
prop(x4, nav, [N]).
prop(x4, det, def).
\end{verbatim}
\end{minipage}

\caption{Prolog representation of the second structure in Fig. \ref{screen}, which is generated from the object representation in Fig. \ref{uml-tree} and used to create the linguistic expression \emph{The old farmers had given soft ducklings to the young women} in the grammar module (cf. section \ref{lexical})} \label{prolog-ucs}
\end{figure}

\subsection{Structures in Functional Discourse Grammar} \label{fdg}

To evaluate the flexibility of our approach, we implemented grammars for structures on the Representational Level (RL) and the Interpersonal Level (IL) in Functional Discourse Grammar (FDG, \citealt{HengeveldAndMackenzie2006}), the successor theory of FG. Fig. \ref{antlr-rl-parser} shows the grammar for structures on the RL, from which a parser is generated that can parse expressions like the structure in Fig. \ref{fdg-rl} into a structure as in Fig. \ref{antlr-tree}. 

\begin{figure}
\begin{footnotesize}
\begin{verbatim}
grammar Representational;

content    : '(' OPERATOR? 'p' X ( ':' head '(' 'p' X ')' )* ')' FUNCTION? ;
soaffairs  : '(' OPERATOR? 'e' X ( ':' head '(' 'e' X ')' )* ')' FUNCTION? ;	
property   : '(' OPERATOR? 'f' X ( ':' head '(' 'f' X ')' )* ')' FUNCTION? ;	
individual : '(' OPERATOR? 'x' X ( ':' head '(' 'x' X ')' )* ')' FUNCTION? ;		
location   : '(' OPERATOR? 'l' X ( ':' head '(' 'l' X ')' )* ')' FUNCTION? ;	
time       : '(' OPERATOR? 't' X ( ':' head '(' 't' X ')' )* ')' FUNCTION? ;

head       : LEMMA? ( '[' 
           ( soaffairs 
           | property 
           | individual 
           | location 
           | time )* ']' ) ? ;

FUNCTION   : 'Ag' 
           | 'Pat'
           | 'Inst' ; //etc.

OPERATOR   : 'Past' 
           | 'Pres' ; //etc.

LEMMA      : 'a'..'z'+ ; 
X          : '0'..'9'+ ;
\end{verbatim}
\end{footnotesize}
\caption{Complete ANTLR v3 grammar for structures on the Representational Level in Functional Discourse Grammar, which describe nested structures as in Fig. \ref{fdg-rl}: each \emph{head} element can take different forms (\emph{content, soaffairs, property, individual, location, time}), which themselves contain a \emph{head} element again} \label{antlr-rl-parser}
\end{figure}

\begin{figure}
\begin{center}
\begin{verbatim}
(p1:[ 
    (Past e1:[
        (f1:tek[
            (x1:im(x1))Ag
            (x2:naif(x2))Inst
        ](f1))
        (f2:kot[
            (x1:im(x1))Ag
            (x3:mi(x3))Pat
        ](f2))
    ](e1))
](p1))
\end{verbatim}
\caption{Underlying structure of a serial verb construction in Jamaican Creole (for \emph{im tek naif kot mi}, 'He cut me with a knife', \citealt[290]{Patrick2004}) on the Representational Level in Functional Discourse Grammar, which is parsable by the parser generated from the rules in Fig. \ref{antlr-rl-parser}. This representation is based on our analysis of the serial verb construction as a single event, which can be backed by native speaker intuition and semantic analysis \citep[291]{Durie1997}; an analysis of a serial verb construction with two events as given in Example 2 of \citet{VanStaden2006} can also be represented using the domain-specific language, while variations in the formal structure would be recognized as invalid}\label{fdg-rl}
\end{center}
\end{figure}

\begin{figure}
\begin{center}
\includegraphics[width=12cm]{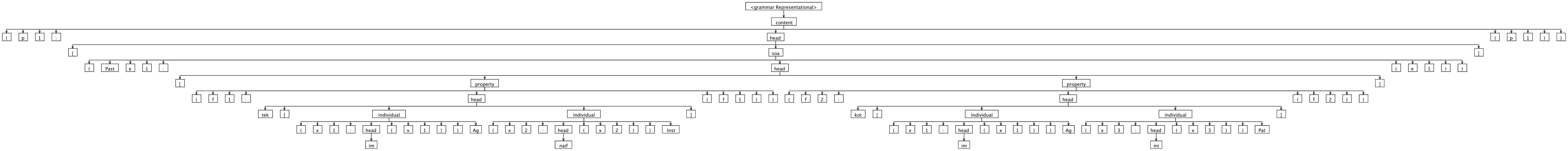}
\end{center}
\caption{Part of the parse tree the parser generated from the rules in Fig. \ref{antlr-rl-parser} produces for the structure in Fig. \ref{fdg-rl}} \label{antlr-tree}
\end{figure}

An ANTLR grammar definition like this provides a validator for the formal structure of RL representations and can be used with a tool like ANTLRWorks (\url{http://www.antlr.org/works/}) to analyse these representations. Having an internal representation of the input (cf. Fig. \ref{antlr-tree}), alternative processing to the creation of the corresponding linguistic expression (as have described for FG structures in section \ref{fg}) is feasible, like output of typeset representations of underlying structures in different formats. This would allow the representation used for publication (e.g. with subscript index numbers, with or without indentation, etc.) to be created from the formal, validated representation.

\subsection{Lexical Entries}\label{lexical}

In the \emph{grammar module} the Prolog representation of the input generated by the \emph{processing module} (cf. Fig. \ref{prolog-ucs}) is used to generate a linguistic expression. Prolog offers convenient notation and processing mechanisms, e.g. lexical entries can be stored directly as Prolog facts (cf. Fig. \ref{prolog-2}). Prolog also has a particular strong standing as an implementation language for FG (e.g. \citealt{Connolly1986,Samuelsdorff1989,Dik1992}). By restricting the usage of Prolog to the grammar module and combining\footnote{For calling Prolog from Java we use Interprolog (\url{http://www.declarativa.com/interprolog/}). The Prolog implementation we use is SWI-Prolog (\url{http://www.swi-prolog.org/}).} it with other languages, instead of using it as a general-purpose programming language for the entire program, we use Prolog as a DSL in one of its original domains.

\begin{figure}
\begin{minipage}[hbt]{6cm}
	\centering
\begin{verbatim}
verb(
   believe,
   state,
   [regular, regular],
   [
       [experiencer, human, X1],
       [goal, proposition, X2]

   ],
   Sat
).
\end{verbatim}
\end{minipage}
\hfill
\begin{minipage}[hbt]{6cm}
	\centering
\begin{verbatim}
verb(
    give, 
    action, 
    [gave, given], 
    [
        [agent, animate, X1], 
        [goal, any, X2],
        [recipient, animate, X3]
    ], 
    Sat
).
\end{verbatim}
\end{minipage}
\caption{Transitive and ditransitive verbs as Prolog facts in the lexicon} \label{prolog-2}
\end{figure}

The expression rules and the lexicon are based on a revised and extended version of the implementation described in \cite{Samuelsdorff1989}. To make the implementation work as a module in the described system, the user dialog of the  original version (in which the underlying structure is built step by step) was replaced by the formal representation that is created in the \emph{input module} and converted into a Prolog representation by the \emph{processing module} (cf. section \ref{fg}). This resembles the shift to a top-down organization \citep[668]{HengeveldAndMackenzie2006} in FDG, where the conceptualization is the first step, not the selection of lexical elements, as it was in FG and in the implementation described in \cite{Samuelsdorff1989}.

\section{Conclusion}

We described a modular implementation of a language generation system, representing underlying structures and lexical entries using domain-specific languages (DSL). The system makes use of an input format based on \cite{Dik1997a} and consists of modules implemented in Java, Prolog and ANTLR\footnote{ANTLR allows further processing in different target languages including Java, C, C++, C\#, Objective-C, Python and Ruby.}. As a first result, this shows that a DSL can be used as a very flexible linguistic expert front-end to a knowledge base in a different language (as we have shown in section \ref{fg} for underlying clause structures based on Functional Grammar that use a Prolog knowledge base). We believe this is a promising way how domain-specific linguistic knowledge can be applied in a natural language processing system.

As all structures in FG and FDG, as well as the lexical entries (which are Prolog facts in our system) have a common tree structure, a unified implementation using ANTLR to define and process all these structures in the same manner as implemented and described for RL representations is feasible. So as a second result, this shows that the concept of a DSL is flexible enough to be applied for newer developments in linguistic theory (as we have shown for structures on the Representational Level in Functional Discourse Grammar in section \ref{fdg}) as well as for extensions of these (as we have shown for structures describing lexical entries in section \ref{lexical}). Therefore domain-specific languages can be used as a tool for consistent formal notation in linguistic description. In our view this encourages the implementation of a full set of grammars for all the structures a linguist creates in linguistic description, which could be the core of software tools that would allow a linguist to create linguistic representations like a programmer writes code, a mathematician writes formulas or a musician writes notes: as something that can actually be validated and even executed in a reproducible manner.

\bibliography{fg-dsl}

\end{document}